\documentclass{article}
\usepackage{ijcai24}

\usepackage{times}
\usepackage{soul}
\usepackage{url}
\usepackage[hidelinks]{hyperref}
\usepackage[utf8]{inputenc}
\usepackage[small]{caption}
\usepackage{graphicx}
\usepackage{amsmath}
\usepackage{amsthm}
\usepackage{booktabs}
\usepackage{algorithm}
\usepackage{algorithmic}
\usepackage[switch]{lineno}
\usepackage[english]{babel}
\usepackage[backend=biber, style=numeric]{biblatex}

\title{MDS-ViTNet: Improving saliency prediction for Eye-Tracking with Vision Transformer}

\author{
Polezhaev Ignat$^{12}$\and
Goncharenko Igor$^{12}$\and
Iurina Natalya$^1$\\
\affiliations
$^1$\emph{Sberbank, Moscow, Russia}\\
$^2$\emph{MIPT, Dolgoprudny, Russia}\\
\emails
polezhaev.im@phystech.edu, ig.goncharenko@gmail.com
}

\begin{document}

\maketitle

\begin{abstract}
    In this paper, we present a novel methodology we call MDS-ViTNet (Multi Decoder Saliency by Vision Transformer Network) for enhancing visual saliency prediction or eye-tracking. This approach holds significant potential for diverse fields, including marketing, medicine, robotics, and retail. We propose a network architecture that leverages the Vision Transformer, moving beyond the conventional ImageNet backbone. The framework adopts an encoder-decoder structure, with the encoder utilizing a Swin transformer to efficiently embed most important features. This process involves a Transfer Learning method, wherein layers from the Vision Transformer are converted by the Encoder Transformer and seamlessly integrated into a CNN Decoder. This methodology ensures minimal information loss from the original input image. The decoder employs a multi-decoding technique, utilizing dual decoders to generate two distinct attention maps. These maps are subsequently combined into a singular output via an additional CNN model. Our trained model MDS-ViTNet achieves state-of-the-art results across several benchmarks. Committed to fostering further collaboration, we intend to make our code, models, and datasets accessible to the public.
\end{abstract}

\section{Introduction}

The human visual system is remarkably complex. This paper concentrates on the mechanisms underlying visual attention, predicting how individuals direct their attention toward specific objects or regions within an image.  There are a number of studies in this field [1, 9-11, 22]. When a person looks at an image, the foveal vision provides the highest resolution visual information. Simultaneously, the peripheral vision, although less detailed, continues to provide long-range visual insights. The eye movements recorded in the experiments serve as the ground truth for estimating image saliency. This paper aims to introduce a novel computer vision model that assigns a saliency value to each pixel of an image, aligning more closely with the ground truth.

The task of predicting human saliency or visual attention maps has widespread applications in numerous fields, such as marketing (for estimating the amount of information a person assimilates while watching an advertisement), robotics, and healthcare.

\section{Related work}

Currently, a variety of methods exist for predicting visual attention maps, encompassing both neural network-based models and traditional approaches. Traditional methods [7, 8, 12, 24] rely on low-level features like luminance, color, texture, and contrast. In contrast, solutions involving neural networks leverage more complex, higher-level patterns.

Many computer vision models, including eye-tracking, employ convolutional neural networks (CNNs) [13, 14, 17, 21]. The success of visual saliency prediction models can be attributed in part to the use of CNNs and to the emerged of large datasets advancing the research in this field. However, CNN-based models have their limitations. Each convolutional kernel in a CNN receives information only from a local subset of image pixels, which restricts these models' ability to capture contextual information over a long distances. Models based on Long Short-Term Memory (LSTM)[5, 18] networks are also utilized. This approach has proven to be effective in processing local and long-range visual information, thereby enhancing the accuracy of saliency prediction.

To tackle the issues mentioned earlier, the TranSalNet model [20], which integrates Transformers into a CNN-based architecture, was introduced in 2022. Transformer encoders learn spatial dependencies across long distances via a multi-head self-attention mechanism. This model has proven to be a state-of-the-art solution on the largest open-source dataset, SALICON. TranSalNet incorporates a backbone formed by convolutional ResNet/DenseNet networks. The CNN generates three feature maps of different dimensions, which are inserted into the transformer encoders followed by a single decoder, producing visual salience maps.

After TranSalNet model, the value of transformers for the task of visual saliency prediction became apparent. The initial discussions about visual transformers emerged in 2020 [26], challenging the convolutional network paradigm with a new architecture that surpassed its convolutional counterparts in accuracy while using fewer parameters. Transformers made a significant impact in the field of computer vision in 2021, with the introduction of the ViT (Vision Transformer) [6].

Additionally, 2021 saw a shift in public focus towards the Swin Transformer models. The Swin (Shifted Windows) Transformer [19], a versatile transformer tailored for computer vision tasks (meaning it's suitable for various applications like semantic segmentation and image classification), outperformed ViT-based and CNN-based architectures. The model authors made new architecture more computationally optimal by utilizing localize attention mechanism. The Shifted Window Multi-Head Attention maintained the network's representative capacity, enabling it to compete with contemporary models. This approach facilitated the development of an architecture that extracts features from images at different spatial scales, allowing Swin Transformer to excel as a backbone in segmentation and detection tasks, where Transformers had previously been less effective. 

\begin{figure*}[h!]
\begin{center}
\includegraphics[width=\linewidth]{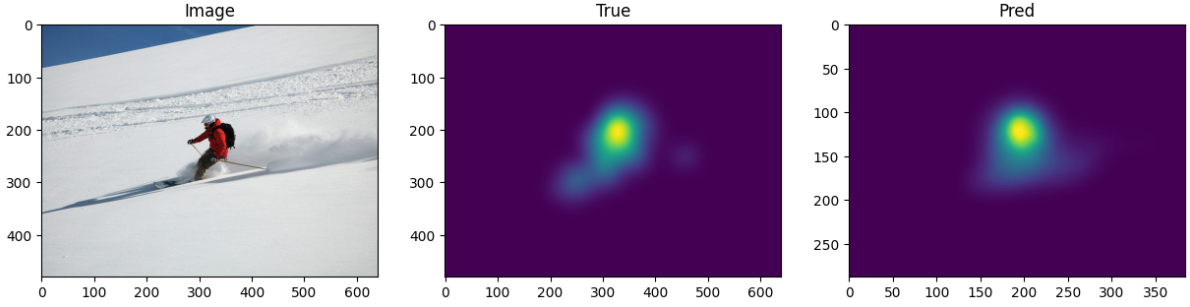}
\caption{Example of model work on an image from SALICON}
\end{center}
\end{figure*}

Contributions of this paper are as follows: (i) substitution of the CNN backbone with visual Swin transformer in the saliency prediction model, (ii) incorporation of multi-decoder architecture with additional CNN-based decoder merging layer, thereby improving the accuracy of the model.

As a result of this work, multi-decoder saliency with visual transformer network (MDS-ViTNet) is proposed and trained on existing datasets. This model demonstrated excellent performance, establishing a new state-of-the-art (SoTA) benchmark. 

The code of the project can be found on the GitHub: https://github.com/IgnatPolezhaev/MDS-ViTNet.

\section{Model Architecture}

\subsection{The ViT encoder}

Extensive research over the years has demonstrated the efficacy of utilizing CNN networks for feature extraction in predicting attention maps. For instance, TranSalNet [20] has achieved SoTA performance in this domain. In this study, two models were developed, employing ResNet-50 and DenseNet-161 as their backbones. Recognizing that TranSalNet employs CNNs as a backbone for image element recognition and Transformer encoders for identifying elements typically detected by human peripheral vision, we were motivated to investigate the potential of Swin Transformers as an alternative backbone to DenseNet and ResNet. This exploration yielded positive results.

In this paper, we introduce MDS-ViTNet (Multi Decoder Saliency by Vision Transformer Network), a novel model that combines transformer components with Swin transformers to capture long-range contextual visual information, a task conventionally handled by CNNs. We utilized a Swin-T transformer from the PyTorch library, with 28M parameters, as the model's backbone. Our experimental findings indicate that employing Swin transformers as the backbone is key to capturing the maximum number of essential qualitative features in images. When these are combined with other types of transformers as encoders and convolutional decoders, they significantly contribute to the model's saliency prediction capabilities, thereby enhancing its perceptual relevance and performance. The unique approach of our proposed model, which leverages a combination of transformers for attention prediction, successfully surpasses the SoTA benchmark in competitions like Salicon.

\subsection{Transformers in the encoder architecture}

Transformers were initially introduced in the context of natural language processing (NLP) [23]. Since then, this architecture has found widespread application in various domains, including time series analysis [25], computer vision [6], and numerous other fields. The popularity of transformers can be attributed to their ability to handle large contexts. Their capacity to retain substantial amounts of information in memory is enabled by the self-attention mechanism. Research has demonstrated their potent capability for capturing long-range information, a feature that can significantly enhance gaze prediction tasks.

Transformers have also gained considerable traction in the field of computer vision. They exhibit competitive performance alongside convolutional networks, which have long been dominant in this area. Currently, there is a diverse array of architectures based on visual Transformers, including DeepViT [28], Swin Transformer [19], and Swin Transformer V2. These models have attained significant success in various visual tasks, predominantly due to the advantages offered by Transformer technology.

In our MDS-ViTNet model, we have implemented similar approaches as in the field of computer vision with Transformers. Our model has a backbone that produces six sets of feature maps of different scales. These feature maps are inserted into the encoder to amplify range and contextual information. The architecture is illustrated in Figure 2. Within this structure, the feature sets $x_2$, $x_4$, $x_6$ with spatial dimensions of (w/8, h/8), (w/16, h/16), (w/32, h/32) are channeled to a first decoder. Meanwhile, the feature sets $x_1$, $x_3$, $x_5$ with spatial dimensions of (w/4, h/4), (w/8, h/8), (w/16, h/16) are channeled to a second decoder. More precisely, the feature sets $x_1$ and $x_2$ undergo a reduction in their third dimension to 512, and the sets $x_3$ to $x_6$ are similarly reduced to 768.

There is no relative or absolute position information in the feature maps, it is necessary to use position embedding (POS) to provide position understanding before feeding input data to the transform encoders. Therefore, as in the work of the TranSalNet colleagues [20], an absolute POS [6] is applied before feeding the input data into the transformer encoders to form a piecewise complement of the input data and a trained matrix of the same shape as the input data. Each transformer encoder contains two identical layers of standard multi-headed self-attention (MSA) and multi-layer perceptron (MLP) blocks [link-link (google 16x16)]. The model employs 8-headed attention in transform coders 1, 2 and 12-headed attention in coders 3 - 6. Layer normalization (LN) and residual coupling are applied before and after each block, respectively. The processing in each transform coder can be represented as follows:

In the absence of relative or absolute positional information within the feature maps, the incorporation of position embedding (POS) is crucial to incorporate positional understanding prior to the input data being fed into the transformer encoders.  Hence, akin to the approach used by authors of [20], we apply an absolute POS [6] before inputting the data into the transformer encoders. This step involves forming a composite of the input data with a trained matrix, which matches the shape of the input data.

Each transformer encoder within our model comprises two identical layers, consisting of standard multi-headed self-attention (MSA) and multi-layer perceptron (MLP) blocks [refer to link-link (google 16x16)]. The model is designed to use 8-headed attention in transformer encoders 1 and 2, and 12-headed attention in encoders 3 to 6. Layer normalization (LN) and residual connections are implemented both before and after each block, respectively.

The processing sequence in each transformer encoder can be described as follows:

\begin{tiny}
    \begin{equation}
        \large z_0 = Conv_{1 \times 1}(x_i) \oplus POS_i, i = 1, 2, 3
    \end{equation}
\end{tiny}%

\begin{tiny}
    \begin{equation}
        \large z_l \textquotesingle = MSA(LN(z_{l-1})) \oplus z_{l-1}, l = 1, 2
    \end{equation}
\end{tiny}%

\begin{tiny}
    \begin{equation}
        \large z_l = MSA(LN(z_l \textquotesingle)) \oplus z_l \textquotesingle, l = 1, 2
    \end{equation}
\end{tiny}%

where $z_l$ are the output feature maps of the l-th layer in the transformer encoder, and $x_i$ are input feature maps from the ViT encoder.

\subsection{Multi-decoder architecture}

Multidecoder architecture (MDA) model in deep learning refers to a mechanism that allows the use of multiple decoders to solve a problem. The advantage of such an architecture is to increase performance and provide faster data processing speed since independent decoders can work in parallel. Also MDA contributes to an error reduction, i.e. if one decoder cannot process the input data correctly, another decoder is able to correct this error. Thus, it is possible to increase the number of parameters by adding new decoders in parallel, which will not increase the inference time as much as adding consecutive layers. MDA is used in many deep learning tasks. For example, in speech technology [29], due to its multiple decoder and counting head architecture, the model requires one forward pass at test time on a single network. MDA has also been mentioned in work on diffusion models [27]. This work breaks the time interval into multiple stages, where specialized multi-decoder U-network architectures that combine time-dependent models with a universal common encoder are used. This approach allows efficient allocation of computational resources and elimination of inter-stage interference, which significantly improves the learning efficiency.

In our work, we propose two independent CNN decoders with one common backbone encoder. The set of features from the transformer-coders (see Fig. 2) fall into the decoders, which predict based on the attention map. Decoder 1 accepts maps $x_2$, $x_4$, and $x_6$, and decoder 2 $x_1$, $x_3$, and $x_5$. The decoders combine these feature maps and restore to the original resolution of the image. Each decoder uses 7 convolution layers. The $x_6$ and $x_5$ maps are inputted into decoders 1 and 2, respectively. After the first and second CNN layers, the significance maps $x_4$ ($x_3$ for decoder 2) and $x_2$ ($x_1$ for decoder 2) are element-wise multiplied with the preceding feature maps. Batch normalization (BN) and activation function (ReLU for all convolution layers except the last one; Sigmoid for the last convolution layer) are applied after each $3 \times 3$ convolution operation ($Conv_{3 \times 3}$), where the former is used to promote convergence and the latter to increase the non-linearity coefficient of the model.

\begin{figure*}[h!]
\begin{center}
\includegraphics[width=\linewidth]{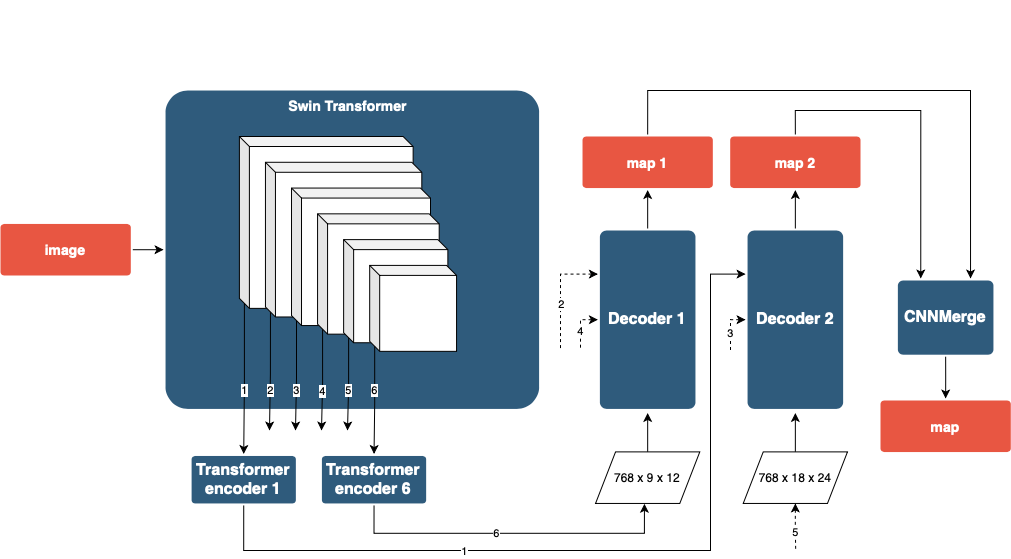}
\caption{MDS-ViTNet architecture}
\end{center}
\end{figure*}

\subsection{An additional model for combining saliency maps}

The multi-decoder architecture in our solution predicts two saliency maps. It was proposed to write an additional CNN-Merge model on convolutional layers that combines these maps and predicts one. CNN-Merge takes as input $x_1$ and $x_2$, where $x_1$ and $x_2$ are two independent importance maps obtained from the decoder. These maps are concatenated by channel dimensionality and processed by convolutional layers. The architecture consists of 7 convolutional layers $Conv_{3 \times 3}$. Batch normalization (BN) and activation function (ReLU) are used after every convolutional layer except the last one. The Sigmoid activation function comes after the 7th layer.

\subsection{Training losses}

The question of how to evaluate a model's ability to predict visual attention map of an image remains an open question. There are many metrics that compare the two saliency maps. The paper [3] gives examples of the latter, and has done a lot of work investigating them. Research in the area of predicting visual saliency maps [4, 15] has shown that using saliency score metrics to define a loss function can significantly improve the model performance.

In our work, following research by TranSalNet colleagues [20], we used a linear combination of different loss functions: Similarity (SIM), Pearson's Correlation Coefficient (CC), Kullback-Leibler divergence (KL). These metrics can be categorized into those that compare the similarity between two distributions and those based on the pixel-by-pixel difference. KL is the metric that baed on thethe pixel-by-pixel difference and the others (SIM and CC) the similarity of the distribution, so in the aggregate loss function KL is taken with a plus sign and the others with a minus sign. In the TranSalNet work, a study was conducted on what coefficients to take in a linear combination of metrics. The minimum loss values on the validation set were fixed, and the coefficients on the loss functions are assigned so that their contribution to the total loss is relatively equal. As a result, weights of 1, 2, and 10 were assigned to SIM, CC, and KL, respectively.

\begin{tiny}
    \begin{equation}
        \large L_{CC} (\hat{m}, m) = \frac{cov(\hat{m}, m)}{\sigma(\hat{m}) \sigma(m)}
    \end{equation}
\end{tiny}%

\begin{tiny}
    \begin{equation}
        \large L_{SIM} (\hat{m}, m) = \sum_i min(\hat{m}_i, m_i)
    \end{equation}
\end{tiny}%

\begin{tiny}
    \begin{equation}
        \large L_{KL} (\hat{m}, m) = \sum_i m_i \log(\varepsilon + \frac{m_i}{\varepsilon + \hat{m}_i})
    \end{equation}
\end{tiny}%

where $\hat{m}$, $m$ - predicted and true saliency maps, $\varepsilon = 2.2204 \times 10^{-16}$ - regularization constant.

\subsection{Evaluation metrics}
In evaluating the quality of visual attention map prediction, various metrics are employed, broadly classified into two categories based on their alignment with human judgments. The first category, 'perception-based metrics', includes measures like CC (Pearson's Correlation Coefficient) and SIM (Similarity). The second category, 'non-perception-based metrics', comprises AUC (Area Under the Curve) and KLD (Kullback-Leibler Divergence). It's important to note that non-perception-based metrics, while not directly measuring perception, are still valuable as they can detect specific gaze behavior patterns, such as topography. Metrics like AUC and KL are particularly suited for detection tasks as they penalize incorrectly identified objects. Others, like SIM, are adept at estimating the relative importance of different regions within an image. In the context of eye-tracking, the AUC metric is commonly used to compare prediction models.

In our study, we developed a custom metric to assess the prediction quality of visual importance maps using AUC. This metric involves selecting a range of threshold values from 0 to 1, corresponding to the range of values in the attention map. For each threshold, True Positives (TP), True Negatives (TN), False Positives (FP), and False Negatives (FN) are calculated point by point. Then, the True Positive Rate (TPR) and False Positive Rate (FPR) for each threshold are computed, leading to the calculation of the area under the curve. This process enables the metric to evaluate the model's predictive ability in terms of accurately identifying true position fixations.

Finally, we compare MDS-ViTNet model with TranSalNet using the metrics SIM, CC, KL, and AUC, providing an evaluation of its performance in predicting saliency maps.

\section{Datasets and data}

\subsection{SALICON and CAT2000 dataset}

In this work, two public datasets, SALICON and CAT2000, were used to train the model. The SALICON  [16] is a large dataset on the popular Microsoft Common Objects in Context (MS COCO) image database. MS COCO is a large-scale image dataset that emphasizes non-iconic images and objects in context. It provides a rich set of specific annotations for image recognition, segmentation and captioning. The SALICON dataset was collected in 2017. It contains 10,000 training images and 5,000 validation images. A set of 5,000 test images without attention maps is also provided.

The CAT2000 dataset [2] contains 2,000 images from 20 different categories with eye tracking data from 24 observers. The entire dataset was not used in the paper, only part of it. This is because some categories are very different from the main distribution of the SALICON dataset, i.e. there were sketches, drawings or monochrome images.

\subsection{Augmentation}

Augmentation was used in this paper to extend the dataset. The augmentation was chosen so that the images do not differ from what a person sees in reality. That is, the image should not be inverted, monochrome or large changes in color. Otherwise, with large and irregular changes, a person's attention will be distorted, for example, in reality some objects cannot be upside down.

Thus, in this paper, augmentation in the form of mirrored horizontal display, small Gaussians blur and changes in contrast, brightness sharpness and saturation were chosen. The images are further normalized and converted to a size of (288, 384).

\section{Experiments}

\subsection{Learning a multi-decoder model}

The model was trained for 15 epochs on an A100 80Gb graphics card. The number of model parameters is 111.5M. 4 images were taken as the batch size. The 8-bit Adam optimizer of bitsandbytes library was used in training. This optimizer was chosen to reduce the video memory consumption in the training in order to increase the size of the batches. StepLR scheduler was also used to reduce the learning rate. There was a stopping criterion in training if 5 epochs in a row loess on validation does not improve the value. The best quality was achieved at the 9th epoch. The results of the model comparison are summarized in the table 1.

\subsection{Learning a CNNMerge model}

The CNN-Merge model was also trained for 15 epochs on an A100 80Gb graphics card. The number of convolution model parameters is about 400K. The size of the batches is 64. AdamW optimizer with learning rate parameter 1e-5 c scheduler StepLR with learning rate halving every 10 epochs. The CNN-Merge training also utilized the stopping criterion. The best quality was achieved at epoch 30. The results of the model comparison are summarized in the table 1.

\begin{table}
    \centering
    \begin{tabular}{lllll}
        \hline
        Model & AUC $\nearrow$ & KL $\searrow$ & CC $\nearrow$ & SIM $\nearrow$ \\
        \hline
        MDS-ViTNet & \textbf{0.8684} & 0.2171 & 0.8949 & 0.7855 \\
        MDS-ViTNet decoder 1 & 0.8650 & 0.2215 & 0.8911 & 0.7817 \\
        MDS-ViTNet decoder 2 & 0.8676 & \textbf{0.2127} & \textbf{0.8980} & \textbf{0.7887} \\
        TranSalNet DenseNet & 0.8486 & 0.2218 & 0.8875 & 0.7757 \\
        \hline
    \end{tabular}
    \caption{Table of model comparison on validation sample}
    \label{tab:plain}
\end{table}

\section{Conclusion}

The conducted research successfully addressed the eye-tracking problem. A thorough analysis of TranSalNet model was undertaken. This led to a comprehensive training cycle for our model, which included compiling an extensive dataset and implementing several imrpovements of the architecture. A key decision was the adoption of the Swin Transformer as an encoder, which has demonstrated superior performance in computer vision tasks compared to traditional encoders based on convolutional layers. Additionally, a multi-decoder architecture was employed, wherein each decoder independently predicts saliency maps. These maps are further processed by an auxiliary CNN model. This additional processing step contributed to an improvement in the quality of the final output. Finally, this approach enabled us to achieve a state-of-the-art result.

\section*{Acknowledgments}

We are grateful to Mark Potanin for his fruitful comments and suggestions. We also thank our colleagues Pavel Abramov and Andrey Mamontov for helpful discussions.

\nocite{*}
\printbibliography

\end{document}